\documentclass[sigconf,natbib=true,nonacm=true]{acmart}

\DeclareUnicodeCharacter{FF08}{(}
\DeclareUnicodeCharacter{FF09}{)}

\newcommand{\frameworkname}{DebiasRAG}

\usepackage{graphicx}
\usepackage{subcaption}

\usepackage{algorithm}
\usepackage{algpseudocode} 

\usepackage{subcaption} 

\usepackage{booktabs}
\usepackage{tabularx}
\usepackage{multirow}

\usepackage{latexsym}      
\usepackage{amsmath}       
\usepackage{amssymb}       
\usepackage{amsfonts}      
\usepackage{amsthm}        
\usepackage{float}         

\usepackage[utf8]{inputenc} 
\usepackage[T1]{fontenc}    

\usepackage{algorithm}     
\usepackage{algpseudocode} 

\usepackage{hyperref}      
\usepackage{geometry}      

\usepackage{graphicx}      
\usepackage{soul}          

\usepackage{tikz}          
\usetikzlibrary{shapes.geometric} 
\usepackage{CJKutf8}       
\usepackage{xcolor}        
\usepackage{tcolorbox}     

\usepackage{microtype}     
\usepackage{booktabs}      
\usepackage{enumitem}      

\AtBeginDocument{%
  }

\author{Rui Chu, Bingyin Zhao, Thanh Quoc Hung Le, Duy Cao Hoang, Huawei Lin, Ping Li, Weijie Zhao, Khoa D Doan, Yingjie Lao}
\renewcommand{\shortauthors}{Chu et al.}
\setcopyright{none}
\begin{document}

\title{DebiasRAG: A Tuning-Free Path to Fair Generation in Large Language Models through Retrieval-Augmented Generation}



\begin{abstract}
Large language models (LLMs) have achieved unprecedented success due to their exceptional generative capabilities. However, they also suffer from producing hallucinations and unexpected outputs (e.g., stereotypes and societal-biased content) .
Prior studies employ LLM fine-tuning and prompt engineering to mitigate such biases in LLMs, which requires additional training or domain knowledge to design the framework. 
However, these approaches either demand extensive training resources or risk degrading the LLM’s original capabilities, while overlooking the need to provide dynamic \emph{debiasing contexts} for fairer inferences. 
In this paper, we propose \frameworkname, a novel LLM-tuning-free and dynamic query-specific debiasing framework based on retrieval augmented generation (RAG) that greatly improves the fairness while simultaneously preserving the intrinsic properties of LLMs. \frameworkname consists of three stages: (1) query-specific debias candidates production; (2) Context candidate pool forming; and (3) Gradient updated debiasing-guided context piece reranking. \frameworkname \space firstly leverages the self-diagnosed \emph{bias contexts} in relevant to the query through regular retrieval. Given the query-specific \emph{bias contexts}, reverse produce \emph{debias contexts}, which will be provided as an additional fairness constraint for LLM output. Secondly, a regular RAG retrieval process will produce the \emph{query-related contexts} from the regular RAG document database. 
During the RAG reranking progress for final debiasing context selection, \frameworkname~re-rank the vectored pieces in the pool based on a gradient-optimized debiasing score as the most \emph{unbiased} support for the user input query.
Lastly, we propose two adaptive scenarios of using \frameworkname, including a regular scenario (with additional RAG documents in the framework) and a non-RAG-document debiasing scenario, which can use \frameworkname as a simple debiasing tool as a replacement for fine-tuning. 
We evaluate the performance of \frameworkname~comprehensively on multiple benchmarks, which demonstrates a highly competitive performance of bias mitigation compared to the state-of-the-art tuning-based methods. 
\end{abstract}




\maketitle




\section{Introduction}



\begin{figure*}[htbp]
    \centering
    \resizebox{0.9\textwidth}{!}{
    \includegraphics[width=1\linewidth]{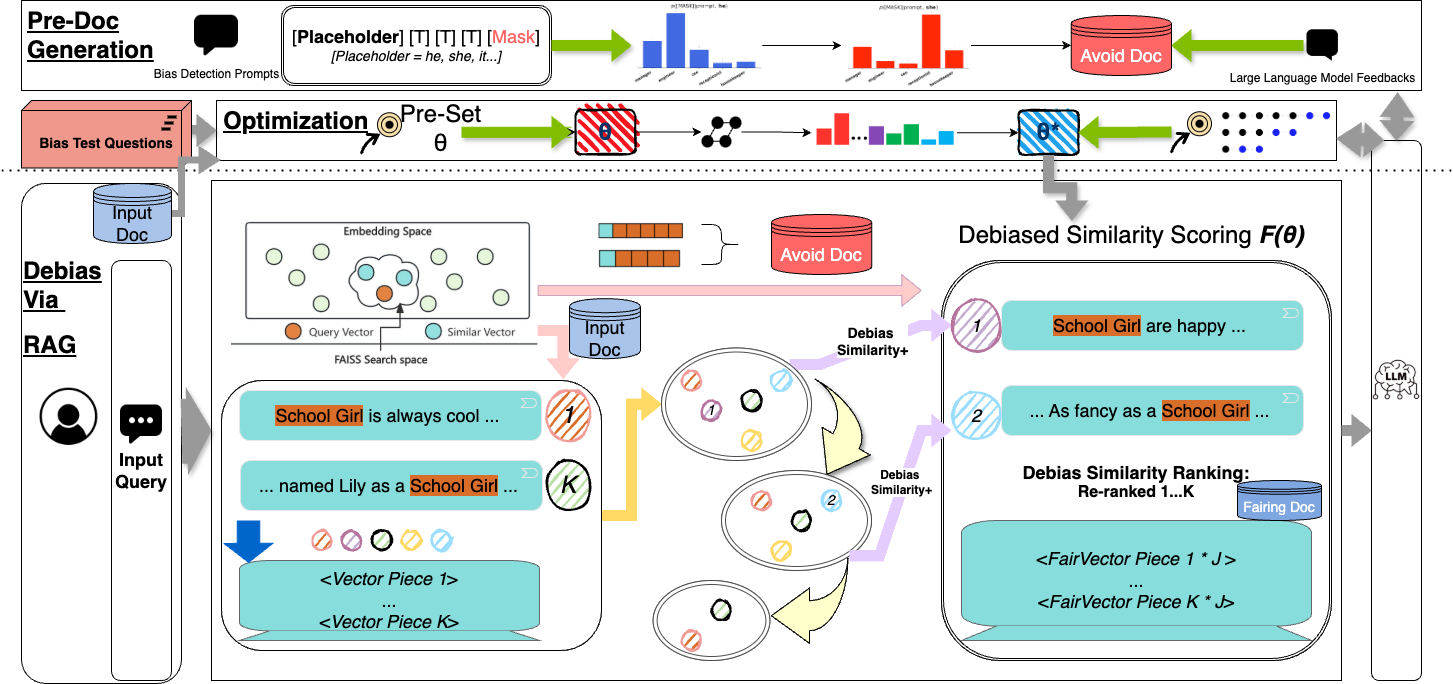}
}
    \caption{System workflow of DebiasRAG. The workflow consists of three main components. The first stage (Upper Block) involves document preparation and preprocessing, including management of the Avoid Document Repo, along with user-provided input documents (Optional). The second stage (Middle Block) performs reverse-generation of debiasing performance based on the user's input to establish a baseline for effective real-time operation. For the third stage (Lower Block), real-time debias-guided reranking optimization, integrates embedding retrieval, gradient-based reranking, and generation, working dynamically to debias the reasoning and output process of large language models.}
    \label{fig:systemFlow}
  
    \vspace{1em}
\end{figure*}

Large language models (LLMs) are widely adopted to empower real-world applications such as content generation~\cite{khalifa2020pcgrl,lin2023magic3d,chen2023fantasia3d, puduppully2019data}, question answering~\cite{zhuang2023toolqa, li2024flexkbqa, louis2024interpretable, wang2024knowledge} and reasoning~\cite{wu2024symbol}. Despite their success, LLMs show inferior performance on domain-specific and knowledge-intensive tasks~\cite{sordoni2024joint,borzunov2024distributed}. A phenomenal example is the hallucinations (e.g., stereotypes and biased information) generated by LLMs when dealing with queries that exceed encapsulated knowledge in the training corpora. 
Prior research~\cite{schick2021self, yang2023adept, hua2023up5, guo2022auto, meade2021empirical},  has revealed that biases are present and can arise at various stages of the nature language processing (NLP) development iterations. Such limitations raise critical concerns, particularly in areas where fairness and equality are highly demanded (e.g., healthcare and legal services) as biased outputs can lead to harmful or even catastrophic consequences~\cite{Zhang2023RAGSurvey,Nazi2023LLMHealthcare}. Thus, it is crucial to mitigate biases that inherently exist in LLMs.

There have been lines of work that attempt to alleviate biases through various approaches. The mainstream mechanisms can be broadly categorized as fine-tuning~\cite{dettmers2024qlora,kaneko2021debiasing} and prompt engineering~\cite{yang2023adept,wang2024prompt, park2023study}. Fine-tuning is a general method to improve LLMs' performance on downstream tasks by retraining the model in an end-to-end manner on a relatively small dataset. In the context of debiasing, one may consider bias elimination as a unique instance of downstream tasks. 
Fine-tuning tends to achieve superior performance since all LLM layers are retrained to accommodate the downstream tasks~\cite{lee2009advances}. 
Yet, it still faces significant limitations, such as high demand for computing resources, the difficulty of data collection, and cumbersome hyperparameter optimization. 
Prompt engineering is an alternative solution that crafts and refines inputs (i.e., prompts) to direct LLMs to generate fair outputs without touching model parameters~\cite{isaev2023scaling}. 
Deliberately designed prompts convey task-specific information that helps models better interpret the input queries and, hence, produce more accurate and reliable outputs. 
However, acquiring useful and effective prompts is challenging, and a common practice usually involves manual design that requires professional expertise and domain knowledge~\cite{ben2022pada,li2024flexkbqa}. 

Another major issue of debiasing in LLMs is the trade-off between fairness and the model's representation ability (e.g., accuracy and expressiveness)~\cite{tsimpoukelli2021multimodal}. 
Prior studies~\cite{yang2023adept, hua2023up5} have shown that an inordinate emphasis on eliminating biases may ultimately undermine the intrinsic properties of LLMs that evolved from pre-training and generate fair yet meaningless outputs.

In this paper, we propose \frameworkname, which aims to address the aforementioned challenges by leveraging retrieval-augmented generation (RAG) to achieve lightweight debiasing comparable to fine-tuning, but without its computational or data overhead. 
Although prompt engineering can also achieve debiasing without parameter updates, it often degrades performance by altering user queries~\cite{guo-etal-2022-auto} or yields limited effects due to its reliance on fixed contexts~\cite{schick2021self}.
In contrast, RAG provides dynamic, query-dependent contexts, but its potential to deliver adaptive debiasing signals—guiding the LLM toward fairer outputs—remains largely unexplored. 
Existing RAG-based debiasing methods primarily focus on retrieving the fairest possible context via embedding-level retrieval guidance~\cite{DBLP:conf/acl/KimSRS25} or agent-based filtering~\cite{DBLP:conf/www/SinghN25}.
While effective in preventing the introduction of new bias, these approaches do not actively reduce existing bias in an otherwise clean LLM.
We instead consider a practical zero-shot scenario in which the debiasing requester possesses only bias-inducing documents the model should learn to avoid, analogous to the fine-tuning setting. 
Unlike prior works that rely on unrelated or neutral corpora (e.g., Wikipedia~\cite{yang2015wikiqa, lewis2020retrieval}) for context substitution or reranking, our approach requires no additional external documents. It is important to note that our goal is to use RAG to debias LLM output (just like a substitution of fine-tuning), which is qualitatively different from \emph{fairness enhancement} within RAG as in prior works~\cite{DBLP:conf/acl/KimSRS25}. 

The overview of the proposed \frameworkname~is shown in Fig.~\ref{fig:systemFlow}. It consists of three main steps: 1) Bias Pre-Doc Generation for RAG, 2) reverse-generation of debiasing piece
based on the retrieved input, and 3) real-time
debias-guided reranking optimization. The key idea is to retrieve query-specific biased contexts and reverse-generate debiasing contexts, which are then integrated into the RAG reranking process. 
The final debiasing context serves as an adaptive augmentation mechanism that constrains the LLM to produce fairer outputs. Our contributions are summarized as follows:



\begin{itemize}[leftmargin=1.3em]
    \item To the best of our knowledge, DebiasRAG is the first framework to demonstrate the effectiveness of RAG for LLM debiasing without requiring additional training or auxiliary data.
    \item We design a debiasing-guided reranking mechanism that steers RAG to retrieve fairness-enhancing contexts, thereby constraining LLM outputs without compromising retrieval quality or generative expressiveness.
    \item We develop a lightweight RAG-based debiasing framework that relies solely on off-the-shelf bias documents, while other RAG documents are optional.
    \item We demonstrate that DebiasRAG achieves a highly competitive performance in bias mitigation compared to state-of-the-art debiasing methods.

\end{itemize}


\section{Related Works} \label{sec:2}


\subsection{Fine-Tuning for Bias Mitigation}
Fine-tuning has achieved impressive advancements in LLM debiasing, which treats debiasing as a downstream task and updates all the model parameters~\cite{zhang2024position,das2024low,agiza2024analyzing}. For example, ~\cite{DBLP:conf/nips/SolaimanD21} mitigates the LLM biases by tuning it on a predefined value-targeted dataset. On the other hand, DPCE (Debiasing Pre-trained Contextualised Embeddings)~\cite{kaneko2021debiasing} enhances the fairness in pre-trained contextualized embeddings by optimizing a debiased embedding loss function to guide the fine-tuning process. Though at the cost of extensive training, the later method achieves a significant fairness gain when debiasing is applied to all tokens and layers of the embedding model and also preserves the model accuracy, which is recognized as the state-of-the-art fine-tuning-based approach. To this end, we consider this approach as one of our baselines. Note that a few relevant earlier works~\cite{kaneko2021debiasing,webster2020measuring} of debiasing techniques were originally focused on static embeddings~\cite{bommasani2020interpreting} while modern LLMs mainly adopt contextualized embeddings.

\subsection{Prompt Engineering for Bias Mitigation}
Prompt engineering is an orthogonal workaround to fine-tuning that eliminates biases by providing instructive descriptions to LLMs. Constructing effective prompts often involves human engagement in the design phase, which imposes a hurdle to users without specific knowledge and domain expertise. For instance, the work in \cite{liu2023disentangled,askell2021general} needed to manually craft a long prompt (e.g., more than 4000 words) from a fictional conversation. 
To address this issue, ADEPT~\cite{yang2023adept} proposes prompt tuning that automates the prompt construction via training exclusively on layers for prompt parameters. As ADEPT outperforms its manually designed counterparts and retains model expressiveness, we utilize it as another baseline. However, ADEPT still needs to train roughly 1\% of LLM parameters, meaning tens or hundreds of millions of parameters are updated at contemporary model sizes.

\subsection{Retrieval Augmented Generation (RAG)}
RAG is an approach that combines LLM with information retrieval techniques, which can help mitigate hallucinations in LLM applications~\cite{lewis2020retrieval,chen2024benchmarking,cheng2024lift}. In RAG, given an input query, the retrieval system first retrieves relevant documents from a large external knowledge base, which are then embedded into vectors and passed to the language model. The language model uses these retrieved documents as additional context to generate more accurate and knowledge-grounded responses, improving performance on tasks that require factual accuracy or specific domain knowledge. Although performing exceptionally in knowledge-intensive and task-specific scenarios and being applied in numerous applications, leveraging RAG in LLM debiasing remains underexplored. In this work, we take the first step to fill this gap.

The issue of social fairness in RAG-retrieved context has recently attracted attention~\cite{DBLP:conf/coling/0002LWT025}. To mitigate bias, retrieval can be guided under fairness constraints, either by steering the process toward fairness-aligned chunks through embeddings~\cite{DBLP:conf/acl/KimSRS25} or by applying agent-based filtering to remove biased contexts~\cite{DBLP:conf/www/SinghN25}. In the computer vision domain, prompt-based filtering to ensure fairness in image generation is well studied~\cite{DBLP:conf/cvpr/ShresthaZCLXD24}. However, fairness-aware reranking in RAG remains underexplored and can draw inspiration from techniques in fair information retrieval~\cite{DBLP:conf/kdd/SinghJ18}.

\vspace{1em}

\section{Methodology}

In this work, we propose a debias-guided retrieval framework \frameworkname~based on RAG that operates with an \emph{avoid-trigger library}, requiring neither additional training nor LLM fine-tuning throughout the workflow nor any architectural modifications to the LLM. Specifically, ~\frameworkname~is designed for reducing the social bias of LLM output under a zero-shot scenario, which requires the \frameworkname~owner to inject the bias documents only.

\subsection{Problem Statement}

The goal of query-adaptive debiasing is to force the targeted LLM $\mathbf{M}$ to generate an output $y_q$ (given user query $q$) under a more bias-mitigated evidence distribution compared to $y_0$, the original output without RAG using only an \emph{avoid set} $\mathbf{A}$, where $\mathbf{A}=\{(\mathbf{a}_i)\}_{i=1}^{n}$ ($i\in[1,n]$) is the social bias piece collection and $\mathbf{a}_i$ is a bias piece. We follow the biased document construction of \emph{Auto-Debias}~\cite{guo2022auto} (detailed in Appendix~\ref{AvoidSet}), but the bias document as an avoid set $\mathbf{A}$ can be changed anytime. 
We consider a scenario where the \frameworkname~owner has access to neither any offline \emph{fair} repository nor the offline training resources, and sometimes there is no any other normal documents in the RAG database (for example, Wikipedia~\cite{yang2015wikiqa}) but only $\mathbf{A}$. 
Let $\mathbf{D}$ be the standard RAG corpus database. The debiasing objective by steps is to,\\
(1) construct a query-specific avoid subset $\mathbf{A}_q\subset\mathbf{A}$,\\
(2) generate a \emph{synthetic} fair set $\tilde{\mathbf{F}}_q$ generated online from $\mathbf{A}_q$, \\
(3) Given the prior results, debias optimize the selection of a top-$K$ context set $C_q$ 
so that using $C_q$ as context constrains the generation $y_q=\mathbf{M}.\mathcal{G}(q;C_q)$ toward a bias-mitigated distribution while preserving original RAG and LLM performance. $\mathcal{G}$ is the LLM generation progress.

\subsection{Method Overview}

We propose \frameworkname~to realize the above objective. 
During the inference, \frameworkname~performs three online steps followed by a debias-guided reranking. \\
First, query-specific avoid retrieval derives $\mathbf{A}_q$ by retrieving the top$K$ pieces from $\mathbf{A}$ as a bias retrieval result of $q$, thereby identifying the bias-relevant themes implicated by the query. \\
Second, online fair synthesis constructs a fair counter-context set solely from $\mathbf{A}_q$ via a counterfactual augmentation generator $g_{\phi}$, following the fairness generation rule of ~\cite{he-etal-2022-mabel, hall-maudslay-etal-2019-name} and NLI context construction~\cite{li2024learning}, producing $\tilde{\mathbf{F}}_q$, the fairness-enhanced contexts.\\
Third, query retrieval over the base corpus obtains a normal candidate set $\mathbf{D}_q$ from $\mathbf{D}$. The candidate pool $\mathcal{C}{}$ is  the union of the normal candidates and the synthesized \emph{query-specific fair subset}..

Given $\mathcal{C}$, \frameworkname~computes a debias-guided score for each candidate that jointly accounts for relevance to $q$ and \emph{farther-from-$\mathbf{A}$} preference with respect to $\mathbf{A}_q$. The final evidence context set $C_q$ is obtained by optimized fairness-reranking for the final Top-$K$ and construct prompt with $q$ as the context delivered to $\mathbf{M}$. 
\frameworkname~therefore uses only the avoid library $\mathbf{A}$ offline ($\mathbf{D}$ is optional) and performs all retrieval, synthesis, and optimized reranking online. 


\subsection{Query-Specific Debias Context}

Given a user query $q$, \frameworkname\ first routes through the avoid set $\mathbf{A}$ (the Bias Documents injected into RAG vector database during Offline) to obtain a query-specific subset $\mathbf{A}_q$. \frameworkname~\emph{retrieves} the top$K$ bias triggers $\mathbf{a}$ based on the highest similarity to the query embedding, detailed in \textit{Line 1 of Algorithm~\ref{alg:debiasrag-routing}}
Using $\mathbf{A}_q$ as conditioning signals, \frameworkname\ then reverse-synthesizes a fair counter-context set.\\
Let $g_{\phi}$ denote a counterfactual augmentation generator.
$\mathcal{V}$ is the token $w$ vocabulary
$\mathbf{a}=(w_1,\ldots,w_T)\in\mathcal{V}^T$, and $T$ is the sequence length of tokens.
$\mathcal{\Omega}$ is a predefined attribute lexicon for debias tokens
(e.g., gender/profession swaps from bias token $w$ to debias token $\tilde w$) with a token-wise substitution map
\[
\qquad
\phi_{\mathcal{\Omega}}(w)=
\begin{cases}
\tilde w,& (w,\tilde w)\in\mathcal{\Omega},\\
w,& \text{otherwise},
\end{cases}
\]
and extend it to sequences by $(\phi_{\mathcal{\Omega}}\circ\mathbf{a})=(\phi_{\mathcal{\Omega}}(w_1),\ldots,\phi_{\mathcal{\Omega}}(w_T))$.
Given an $\mathbf{a}\in\mathbf{A}_q$, 
the generator $g_{\phi}$ produces a counter-bias output by lexicon-guided paired substitutions:
\[
g_{\phi}(\mathbf{a},q) \;=\; \mathrm{Refine}\Big(\phi_{\mathcal{\Omega}}\circ\mathbf{a}\\;q\ \Big),
\]
where $g_{\phi}$ maps sensitive terms through a predefined attribute lexicon $\mathcal{\Omega}$, 
and $\mathrm{Refine}(\cdot)$ applies a lightweight rewrite to ensure fluency by perplexity~\cite{jelinek1977perplexity}, generate a rewritten counterfactual candidate and keep the most fluent one under a perplexity filter~\cite{kaneko2021debiasing}. 
The query-specific fair subset $\tilde{\mathbf{F}}_q$ is produced as shown in \textit{Line 2 of Algorithm~\ref{alg:debiasrag-routing}}


\textbf{In parallel, \frameworkname\ performs \emph{standard retrieval}} over the base RAG corpus $\mathbf{D}$ (for example, the Wikipedia~\cite{yang2015wikiqa}) to form a query-specific normal candidate set $\mathbf{D}_q$ formed as shown in \textit{Line 3 of Algorithm~\ref{alg:debiasrag-routing}},
The candidate pool $\mathcal{C}$ is
\begin{equation}
\mathcal{C} \;=\; \mathbf{D}_q \,\cup\, \tilde{\mathbf{F}}_q,
\label{eq:pool}
\end{equation}
We use \emph{FAISS}~\cite{douze2024faiss} for $\mathrm{sim}(\cdot,\cdot)$, detailed in Appendix~\ref{Additional Settings}.

\begin{algorithm}[t]
\caption{Query-Specific Fair Context Synthesis}
\label{alg:debiasrag-routing}
\begin{algorithmic}[1]
\Require query $q$; avoid library $\mathbf{A}$; base corpus $\mathbf{D}$; generator $g_{\phi}$; integers $k$
\State $\mathbf{A}_q \leftarrow \texttt{Top}_{K}\big(\mathrm{sim}(q,\mathbf{a})\big),\ \mathbf{a}\in\mathbf{A}$
\State $\tilde{\mathbf{F}}_q \leftarrow \{\, g_{\phi}(\mathbf{a},\mathrm{topic}(q)) \mid \mathbf{a}\in\mathbf{A}_q \,\}$
\State $\mathbf{D}_q \leftarrow \texttt{Top}_{K}\big(\mathrm{sim}(q,\mathbf{d})\big),\ \mathbf{d}\in\mathbf{D}$
\State $\mathcal{C} \leftarrow \mathbf{D}_q \cup \tilde{\mathbf{F}}_q$
\State \Return $\mathcal{C}$
\end{algorithmic}
\end{algorithm}

\subsection{Debias-Guided Reranking}

\frameworkname\ assigns each debias candidate $c\in\mathcal{C}$ a \emph{debias-guided score} that balances query relevance and a distance-from-avoid (larger is better) preference:
{uery-relevance score} $s_q(c)$ of candidate $c$, as shown in \textit{Line 1 of Algorithm~\ref{alg:debiasrag-rerank-online}} 
and {distance-from-avoid score} $s_a(c)$ of candidate $c$ as shown in \textit{Line 2 of Algorithm~\ref{alg:debiasrag-rerank-online}}

\subsubsection{Candidate filtering.}
To avoid retaining contexts that are still too close to the retrieved avoid triggers, we discard candidates with high avoid-similarity:
\begin{equation}
\mathcal{C} \leftarrow \Big\{\, c \in \mathcal{C} \;:\; \max_{a \in A_q} \mathrm{sim}(c,a) \le \tau_{\text{avoid}} \,\Big\}.
\label{eq:candidate-filter}
\end{equation}

The hyperparameters are shown in the implementation details.

\subsubsection{Linear \emph{Debias Scoring} over per-pool normalized}

Although both signals are normalized per query, their discriminative power varies across candidate pools: some queries yield nearly constant relevance but highly variable avoid-distance, while others exhibit the opposite.

The debiasing Score $S$ can be calculated by $\widetilde{\cdot}$, the scores after per-pool $C$ normalization to $[0,1]$ (as described in \textit{Line 5 of Algorithm~\ref{alg:debiasrag-rerank-online}}, detailed in Appendix.~\ref{Normalization}) with parameter $\theta$,
\begin{equation}
S_\theta(c\mid q)=\theta^\top \phi_q(c),\qquad \theta\in\Delta_2.
\label{eq:score-linear}
\end{equation}
where \[
\phi_q(c)=\begin{bmatrix}\widetilde{s}_q(c)\\ \widetilde{s}_a(c)\end{bmatrix},\quad
\Delta_2=\{\theta\in\mathbb{R}^2_{\ge 0}:\ \|\theta\|_1=1\},\quad
\]

\subsubsection{Online Gradient Update for better reranking}
In order to have the best performance of \emph{debias-guided reranking}, \frameworkname~update $\theta$ by gradient online, as described in \textit{Line 7 of Algorithm~\ref{alg:debiasrag-rerank-online}}.
Let \emph{Utility} for target,
\begin{equation}
u(c)=\widetilde{s}_q(c)+\widetilde{s}_a(c),\qquad
\hat u(c)=\frac{u(c)-\min_{c'\in\mathcal{C}}u(c')}
               {\max_{c'\in\mathcal{C}}u(c')-\min_{c'\in\mathcal{C}}u(c')},
\label{eq:u-hatu-mm}
\end{equation}
and define the target
\begin{equation}
y_{q,c}=\frac{\exp(\hat u(c))}{\sum_{c'\in\mathcal{C}}\exp(\hat u(c'))},\qquad
\label{eq:y-target-nolambda}
\end{equation}
\noindent\textit{where} $c'$ is a {dummy index} ranging over the pool $C$.
With the model-induced list distribution
\begin{equation}
p_\theta(c\mid q)\;=\;\mathrm{softmax}_c\!\big(S_\theta(c\mid q)\big)
\;=\;\frac{\exp\!\big(S_\theta(c\mid q)\big)}{\sum_{c'\in\mathcal{C}}\exp\!\big(S_\theta(c'\mid q)\big)}.
\label{eq:softmax-score}
\end{equation}
The listwise cross-entropy loss~\cite{DBLP:conf/icml/XiaLWZL08} is
\begin{equation}
\mathcal{L}_{\text{}}(\theta;q)
=-\sum_{c\in\mathcal{C}} y_{q,c}\,\log p_\theta(c\mid q),
\label{eq:listwise-nolambda}
\end{equation}
encouraging $p_{\theta}$ to match the target $y_{q,c}$.
We take one streaming step
\begin{equation}
\theta \leftarrow \Pi_{\Delta_2}\!\Big(\theta-\eta\,\nabla_\theta \mathcal{L}_{\text{}}(\theta;q)\Big),\qquad
\label{eq:streaming-step-nolambda}
\end{equation}

\noindent\textit{where} $\eta>0$ is the stepsize,
$\nabla_\theta \mathcal{L}_{\text{}}(\theta;q)$ is the gradient of the listwise loss for query $q$,
and $\Pi_{\Delta_2}(\cdot)$ is the projection onto the 2-simplex.\\
The final \frameworkname ing context set is $C_q$.

\begin{algorithm}[t]
\caption{Debias-Guided Reranking with Online Streaming Update}
\label{alg:debiasrag-rerank-online}
\begin{algorithmic}[1]
\Require query $q$; pool $\mathcal{C}$; query-specific avoid set $\mathbf{A}_q$; scorer params $\theta$; integer $K$
\For{$c \in \mathcal{C}$}
    \State $s_q(c) \leftarrow \mathrm{sim}(q,c)$
    \State $s_a(c) \leftarrow 1 - \max_{\mathbf{a}\in\mathbf{A}_q}\mathrm{sim}(c,\mathbf{a})$
\State $\widetilde{s}_q(\cdot),\widetilde{s}_a(\cdot) \leftarrow \texttt{NormalizeOver}(\mathcal{C})$
    \State $\theta \leftarrow \texttt{Optimize}\big(\theta;\ q,\ \mathcal{C}\big)$
    \State \emph{Calculate} $S_\theta(c\mid q)$
\EndFor
\State $C_q \leftarrow \texttt{TopK}_{\,c\in\mathcal{C}}\, S_\theta(c\mid q)$
\State \Return $C_q$
\end{algorithmic}
\end{algorithm}

The step size $\eta$ is selected by backtracking line search to ensure
$\mathcal{L}(\theta^{t+1};q)\le \mathcal{L}(\theta^{t};q)$.

\section{Experiments}

\subsection{Experimental Settings}
We conduct the experiments on multiple NVIDIA H100 GPUs and conduct performance evaluations on the pretrained OPT family model~\cite{zhang2022opt}, BB3 family model~\cite{shuster2022blenderbot}, LLaMa model family~\cite{touvron2023llama}, and GPT2~\cite{radford2019language} (more in Appendix). We use BERT(SBERT) for embedding and FAISS~\cite{douze2024faiss} for similarity search, respectively, and more RAG architectures will be compared..

\subsection{Benchmarks and Metrics}

\subsubsection{Benchmarks.} We follow the same practice in existing works~\cite{kaneko2021debiasing,yang2023adept,guo2022auto} , and evaluate \frameworkname's \space debiasing performance and expressiveness on standard bias benchmarks, including StereoSet~\cite{Nadeem2020StereoSetMS}, CrowS-Pairs~\cite{nangia-etal-2020-crows} , and SEAT~\cite{may2019measuring}. \textbf{StereoSet} offers a multifaceted assessment of LLMs' bias and their ability to produce meaningful semantic information via a cloze test, which requires LLMs to choose the best answer from three options (i.e., a stereotype, an anti-stereotype, and an unrelated) with a given context. \textbf{CrowS-Pairs} consists of paired test sentences that differ only in a stereotyped or anti-stereotyped word in the same position. This benchmark evaluates whether a language model assigns a higher probability to stereotyped sentences compared to their anti-stereotyped counterparts while attempting to account for differing priors. 


\subsubsection{Metrics} To comprehensively and extensively evaluate \frameworkname, we adopt a variety of debiasing metrics that are mostly used for aforementioned benchmarks: Language Modeling Score (LMS), Stereotype Score (SS), Idealized CAT Score (ICAT), and CrowS-Pairs Score (CP-S). Among these, LMS, SS, and ICAT come from StereoSet, where we measure one type of bias, i.e., gender, as well as the overall performance.

\begin{itemize}
    \item LMS indicates expressive capability. It ranges from 0 to 100. \ul{A higher value represents better performance.} 
    \item SS mainly measures biases. \ul{A value closer to 50 means less bias.} 
    \item ICAT combines the above two metrics and represents optimal performance in both language modeling and bias mitigation. It has a max value of 100. \ul{A higher value represents better performance.} 
    \item CP-S evaluates the probability of assigning stereotyped or anti-stereotyped sentences on CrowS-Pairs. \ul{A value closer to 50 means less bias.}
    \item SEAT~\cite{may-etal-2019-measuring} tests 6 and 6b assess the association of male and female terms with career and family attributes, while tests 7, 7b, 8, and 8b examine the link between male and female names and career or family-related words.
\end{itemize}

\paragraph{Baseline Methods}
\emph{Prompting}~\cite{esiobu2023robbie}, lightweight prompt engineering for bias reduction;
\emph{Self-Debias}~\cite{schick2021self}, a decoding-time self-debiasing strategy;
Sentence-Debias~\cite{liang-etal-2020-towards}, learns a bias direction from counterfactual sentence pairs and projects it out of the model’s sentence representations.
Adapter-Tune~\cite{DBLP:conf/nips/SolaimanD21}, inserts small bottleneck adapter layers into each Transformer block and fine‑tunes only these adapters with a debiasing objective.

\subsubsection{Pre-Processed documents}
\paragraph{Normal RAG Repository $D_{\text{normal}}$}. The normal RAG repository is a repository that simulates the repo so the user will add their Document as a regular RAG usage. In this paper, the normal repo will be auto-filled with a Pre-Processed Fair-Enhancement Data doc as a base to avoid an empty repo scenario and fill in a mini-Wikipedia dataset~\cite{hewlett2016wikireading} to simulate user input.\\
\paragraph{Avoid Doc Repository $\mathbf{A}$ ($D_{\text{avoid}}$)} Avoiding Doc repository will be filled with a pre-processed Model-self-diagnosed data. In our method, the document is generated by diagnosing Bert.\\
\paragraph{Bias Testing Questions for Optimization}. In order to optimize the parameter $\theta$, we choose Stereoset as our Bias Testing dataset. For each optimization progress, we will randomly pick $\lambda$ questions from Stereoset (where we set $\lambda$ = 3) and maximize either the ICAT score or optimize SS scores.

\subsection{Debiasing Performance}

\subsubsection{Step-By-Step Debias Flow}
\begin{figure}[htbp]
    \centering
    \begin{subfigure}[t]{0.48\columnwidth}
        \centering
        \includegraphics[width=\linewidth]{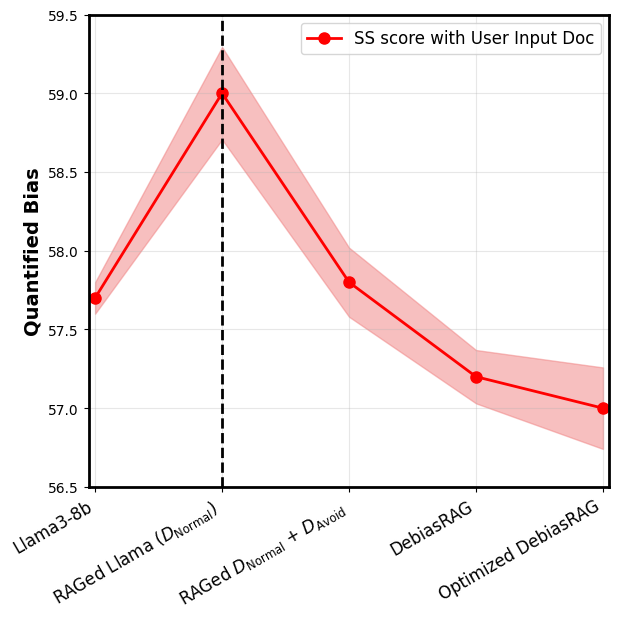}
        \caption{SS score step-by-step optimization considering a scenario of user inputing $\textbf{D}$ first and optimizing $\theta$ next}
        \label{fig:UserRAG}
    \end{subfigure}%
    \hfill
    \begin{subfigure}[t]{0.48\columnwidth}
        \centering
        \includegraphics[width=\linewidth,height=4cm,keepaspectratio]{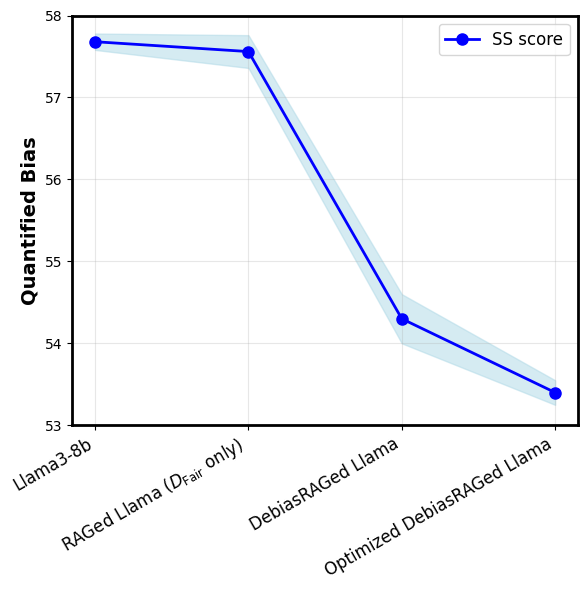}
        \caption{SS score decreasing on a step-by-step optimization with \frameworkname (Inference with only Bias document (Avoid Doc) $\textbf{A}$)}
        \label{fig:FairStereo}
    \end{subfigure}
    \caption{The optimization performance of a same $\lambda$ with the increasement of iterations}
    \label{fig:combined}
\end{figure}

The workflow is shown as a step-by-step optimization result with a biasing score reducing flow in Fig \ref{fig:combined}.
\paragraph{RAG Document-free Scenario} 
Taking the \emph{StereoSet score} benchmark as an example, we compare the performance of 2 scenarios. \\
In this scenario, there is no mini-Wikipedia dataset be added to $\textbf{D}$ ($D_{\text{normal}}$), but remaining a NLI dataset. The optimization flow of the scenario can be viewed as followed Figure \ref{fig:FairStereo}.
In the Figure, we are showing the optimization steps of our methods: the original performance of Llama3-8b, RAGed Llama with only $D_{\text{Fair}}$, \frameworkname ed LlaMa, and Gradient Optimized \frameworkname LlaMa; with increasing performance of Stereoset fairness score ICAT. \\
It can be found that just putting Fairness-enhancement documents into RAG is not very useful for large language models like LlaMa3. However, \frameworkname~can obviously decrease the bias score from the original 57.7 to 54.3 (closer to 50 is the best), and with a quick optimization using a $\lambda$ = 3 to update $\theta$, the performance can be improved more until 53.4. \\

\paragraph{RAG Document available Scenario} Putting mini-Wikipedia dataset into $D_{\text{normal}}$ as a simulation of the scenario that the user has added their own documents they want to use as a traditional RAG. The flow chart can be viewed in Figure~\ref{fig:UserRAG}. 
In the chart, we can see the optimization steps starting from a bias increase caused by the user-input data, which is a mini-Wikipedia dataset $D_{\text{normal}}$. 
We can find that even though the user-input RAG document brings a huge bias during the inference stage, \frameworkname~framework can also handle this scenario, making the quantified Bias drop.
\subsubsection{Optimization Performance}
\begin{figure}[htbp]
    \centering
    \begin{subfigure}[t]{0.48\columnwidth}
        \centering
        \includegraphics[width=\linewidth]{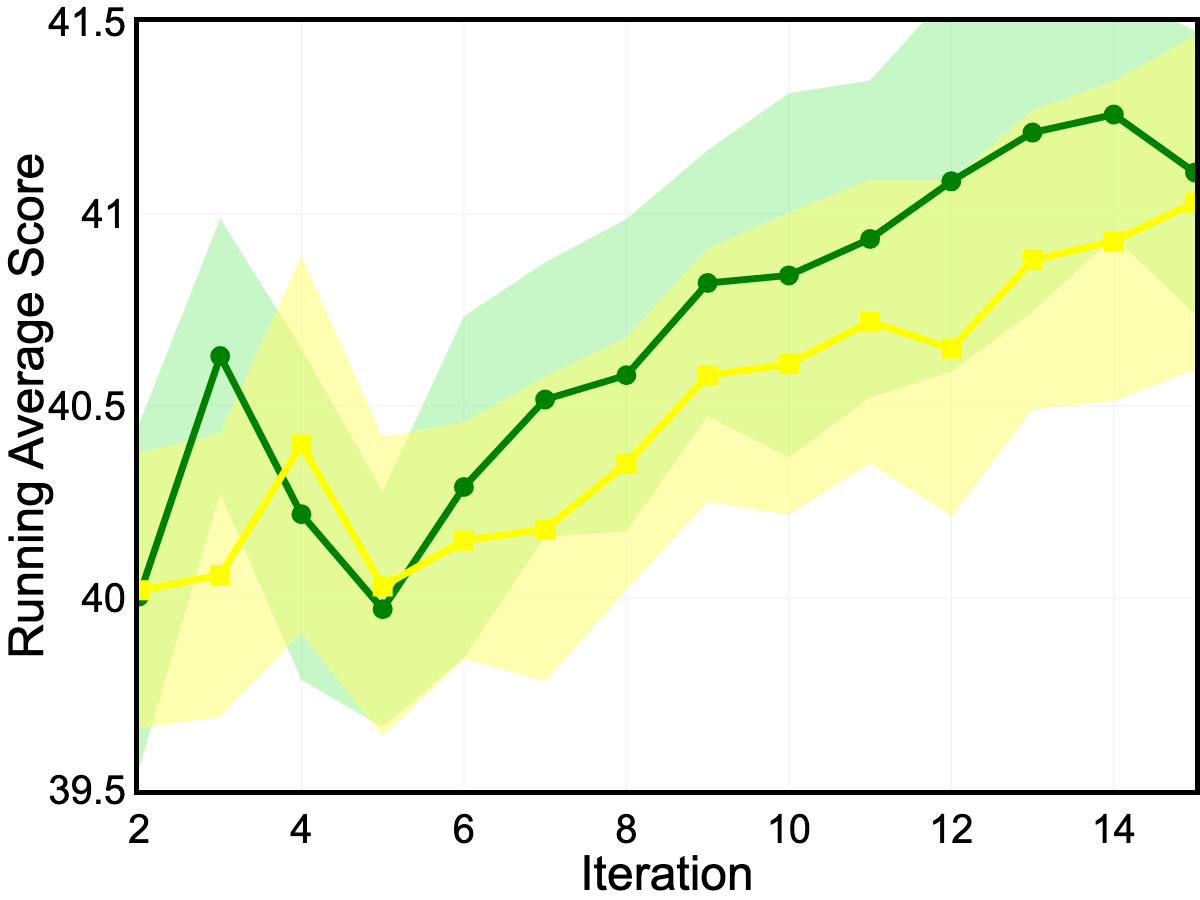}
        \caption{The Green Line shows $\tilde{\mathbf{F}}_q$ performance and the yellow line shows $\textbf{D}$ performance. A higher running score indicates predictions are more balanced (i.e., the deviation from SS 50 is smaller)}
        \label{fig:SameLambdaOptimization}
    \end{subfigure}%
    \hfill
    \begin{subfigure}[t]{0.48\columnwidth}
        \centering
        \includegraphics[width=\linewidth]{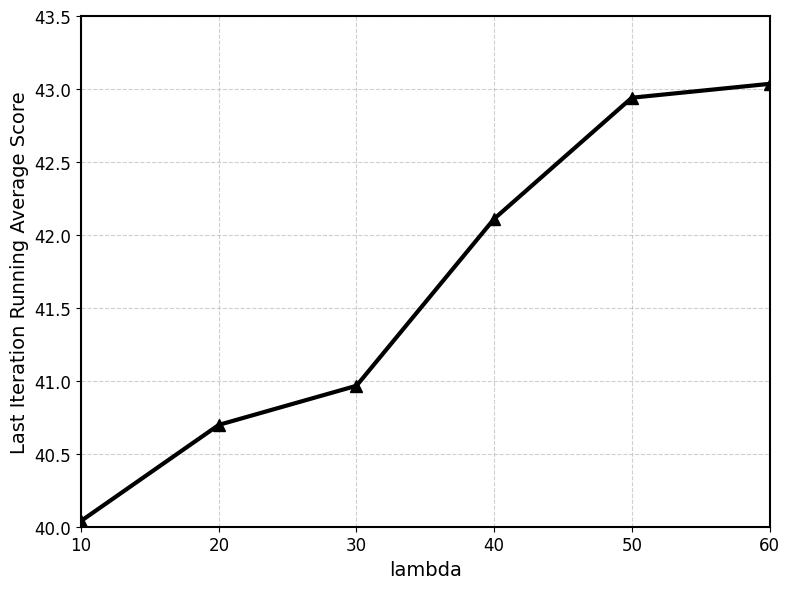}
        \caption{Optimization effectiveness with the Increase of $\lambda$. The Increase of $\lambda$ has coorelated the Increase of optimization performance and thus the $\lambda$ could be used as an optimization factor.}
        \label{fig:LambdaIncrease}
    \end{subfigure}
    \caption{Effectiveness of Optimization}
    \label{fig:CombinedFigures}
\end{figure}

\paragraph{Performance improvements within the same $\lambda$} As it is shown in Figure \ref{fig:SameLambdaOptimization},  the running average of the target scores across iterations provides a clear visual indication that our gradient optimization process is steadily converging towards an optimal solution. This convergence not only demonstrates that the parameter estimates are progressively refined, but also substantiates the overall effectiveness of our optimization strategy. \\
\paragraph{Performance improvements with increasing $\lambda$}  By collecting the same final iteration performance on different $\lambda$, which is shown in Fig \ref{fig:LambdaIncrease}, it can be found that the $\lambda$ can influence the optimization performance and will have a potential coverage point. In our experiment, we find that when the $\lambda$ comes to 60, the performance remains a high standard.


\subsubsection{General Performance}

Tables~\ref{tab:robbie-t6-bias-bb3}, ~\ref{tab:GPTSTEREO1} and \ref{tab:GPTSTEREO} summarize the debiasing performance of \frameworkname~applied to Large Models, such as \emph{BB3-175B}, and GPT2 in comparison with the original GPT2 model. 

\paragraph{LLMs across methods and models.}
Table~\ref{tab:robbie-t6-bias-bb3} shows the debiasing performance compared to prior methods on \emph{state-of-the-art} Language Model (\textbf{BB3-175B}). \frameworkname overperformed both \emph{Self-Debias} and \emph{Prompting} (prompt-level debiasing methods) in most of the benchmarks, showing the outstanding debiasing performance as a prompt-side debiasing approach.

\paragraph{Compared with the clean language model.}
According to Table~\ref{tab:GPTSTEREO1},
in terms of the CrowS-Pairs Score (CP-S), \frameworkname~achieves a score of 41.38, which is an improvement over the original, thereby moving the score marginally to the best point.
For the StereoSet-gender benchmark, the results reveal that the LMS score decreases notably after applying \frameworkname. 
Overall, as for the \emph{StereoSet-Gender}, \frameworkname~achieves better SS and ICAT, showing \frameworkname can maintain LLM capability while achieving a good debias performance.

The overall StereoSet performance benefits substantially from DebiasRAG. 
The overall LMS score increases for around 9\%, 
while the overall SS score decreases dramatically from 57.6 to 49.72 (closer to 50 is better), thus approaching the ideal benchmark. 
Furthermore, the overall ICAT score increases significantly from 70.0 to 90.53, proving the strong debiasing effectiveness of \frameworkname.

In addition, regular RAG was compared as a baseline on a large language model, LLaMa3-8b, as shown in Table~\ref{tab:GPTSTEREO}. Similarly, under StereoSet-Gender, \frameworkname overperforms both Regular RAG and the clean LLM on debiasing.

These shifts suggest that, although the debiasing process affects gender-specific expressiveness to some extent, it is highly effective in mitigating bias on a broader scale.

\begin{table}[t]
\small
\centering
\setlength{\tabcolsep}{3.5pt}
\renewcommand{\arraystretch}{1.0}
\begin{tabular}{lrrrrr}
\toprule
\textbf{Method} & \textbf{BOLD (\%)} & \textbf{TG2 (\%)} & \textbf{APS (\%)} & \textbf{Regard (\%)} & \textbf{HBR (\%)} \\
\midrule
Base           & 19.3 & 29.0 & 34.6 & 29.7 & 79.2 \\
+ Prompting    & 17.7 & 21.3 & 20.0 & 19.5 & 72.1 \\
+ Self-Debias  & 17.9 & 26.0 & 33.1 & 33.0 & 94.8 \\
+ \textbf{\frameworkname~}  & 17.4& 21.2& 20.0& 33.9& 96.7\\
\bottomrule
\end{tabular}
\caption{\textbf{Bias} (\%) for BB3-175B.}
\label{tab:robbie-t6-bias-bb3}
\end{table}

\begin{table}[htbp]
\centering

\setlength{\tabcolsep}{8pt}
\resizebox{\linewidth}{!}{
\begin{tabular}{lcc}
\toprule
\hline
& original& \frameworkname~\\
\midrule
CrowS-Pairs Score (CP-S) & 41.05 ±0.01& 41.38 ±0.01\\
\hline
StereoSet-gender: LMS& 98.76 ±0.02& 92.98 ±0.04\\ 
StereoSet-gender: SS& 54.81 ±0.03& \textbf{51.44} ±0.02\\ 
StereoSet-gender: ICAT& 89.26 ±0.01& 90.21 ±0.06\\ \hline 
StereoSet-overall: LMS& 82.51 ±0.03& 91.05 ±0.04\\  
StereoSet-overall: SS& 57.60 ±0.03& \textbf{49.72} ±0.05\\
StereoSet-overall: ICAT& 70.02 ±0.02& 90.53 ±0.01\\
\hline \bottomrule

\end{tabular}
}
\caption{Debiasing Performance on GPT.}
\label{tab:GPTSTEREO1}
\end{table}



\begin{table*}[htbp]
\vspace{1em}
\centering
\caption{Debiasing Performance on LLaMa3-8b}
\label{tab:GPTSTEREO}
\renewcommand{\arraystretch}{1.0} 
\setlength{\tabcolsep}{16pt}  
\resizebox{1.0\linewidth}{!}{ 
\begin{tabular}{lccc}
\toprule
\hline
& Original& Regular RAG& \frameworkname \\
\midrule
CrowS-Pairs Score (CP-S) & 40.08$\pm$ 0.03& 41.01$\pm$ 0.02& 41.73 $\pm$ 0.05\\
\hline
StereoSet-gender: LMS  & 99.59$\pm$ 0.04& 92.56$\pm$ 0.03& 92.15 $\pm$ 0.08\\ 
StereoSet-gender: SS   & 57.70$\pm$ 0.02& 58.93$\pm$ 0.05& \textbf{53.85} $\pm$ 0.07\\ 
StereoSet-gender: ICAT & 84.30$\pm$ 0.04& 76.03$\pm$ 0.09& \textbf{85.05} $\pm$ 0.19\\
\bottomrule
\end{tabular}
}
\vspace{1em}
\end{table*}

\paragraph{SEAT Metric.}
In Table~\ref{tab:seat_on_bert}, we evaluate our method against SEAT tests, which can provide insights into gender and racial biases. 

We also compare the results to the Dropout debiasing~\cite{webster2020measuring}, which uses dropout regularization to reduce overfitting to bias information (e.g., gender) and thereby achieves debiasing. It can be seen that the DebiasRAG generally shows a reduction in bias, particularly for SEAT-6 and SEAT-8, indicating effective debiasing in these benchmarks. 
\begin{table}
\centering
\setlength{\tabcolsep}{8pt}
\resizebox{\linewidth}{!}{
\begin{tabular}{lccc}
\toprule
\hline
& original &Dropout&DebiasRAG  \\ \hline 
 SEAT-6& 0.48&0.38&0.42\\
 SEAT-6b& 0.11&0.38&0.36\\
 SEAT-7& 0.25&0.31&0.25\\
 SEAT-7b& 0.25&0.40&0.25\\
 SEAT-8& 0.40&0.48&0.35\\
 SEAT-8b& 0.64&0.58&0.64\\ 
 \hline
\bottomrule
\end{tabular}
}
\caption{SEAT Performance}
\label{tab:seat_on_bert}
\vspace{-1em}
\end{table}





\begin{figure}
    \centering
    \includegraphics[width=0.75\linewidth]{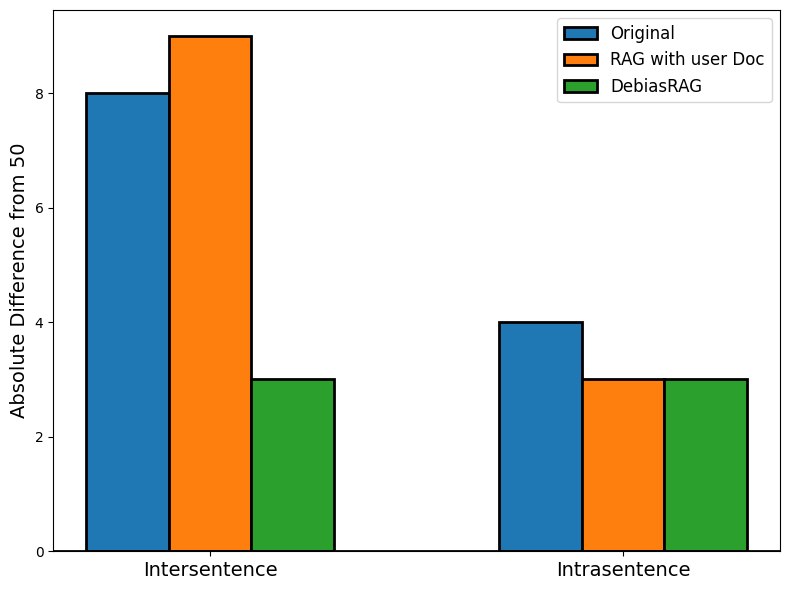}
    \caption{SS score on Llama3-8b with Intersentence and Intrasentence performances}
    \label{fig:langtest}
\end{figure}

\paragraph{Intersentence and Intrasentence Tests}
LangTest~\cite{langtest_stereoset} uses two main tests: intersentence and intrasentence (detailed in Appendix). 
We also conduct this set of experiments on LlaMa3-8b, as shown in Figure~\ref{fig:langtest}.  Under the intersentence criterion, the original model exhibits a bias score that increases slightly to 59 when user-provided documents are incorporated with RAG. 
This suggests that user inputs may introduce additional bias. 
In contrast, \frameworkname~substantially reduces the intersentence bias score to 53, demonstrating its strong capability to mitigate bias in scenarios where external documents tend to elevate the bias level.

For the intrasentence setting, the original model and the RAG with user documents yield scores of 46. 
Although these scores are already close to the ideal benchmark (assumed to be 50), \frameworkname~adjusts the score to a better fairness generation. 
When considering the absolute deviation from the ideal value, the original model deviates by 4 points (i.e., |46–50| = 4) while DebiasRAG deviates by only 3 points (i.e., |53–50| = 3). 
Thus, even in the intrasentence scenario—where the bias is less pronounced—\frameworkname~still achieves a measurable improvement over the original model.


\subsubsection{Comparison with Tuning-based Methods}

\begin{figure}[t]
  \centering
  \begin{subfigure}{\columnwidth}
    \centering
    \includegraphics[width=\columnwidth]{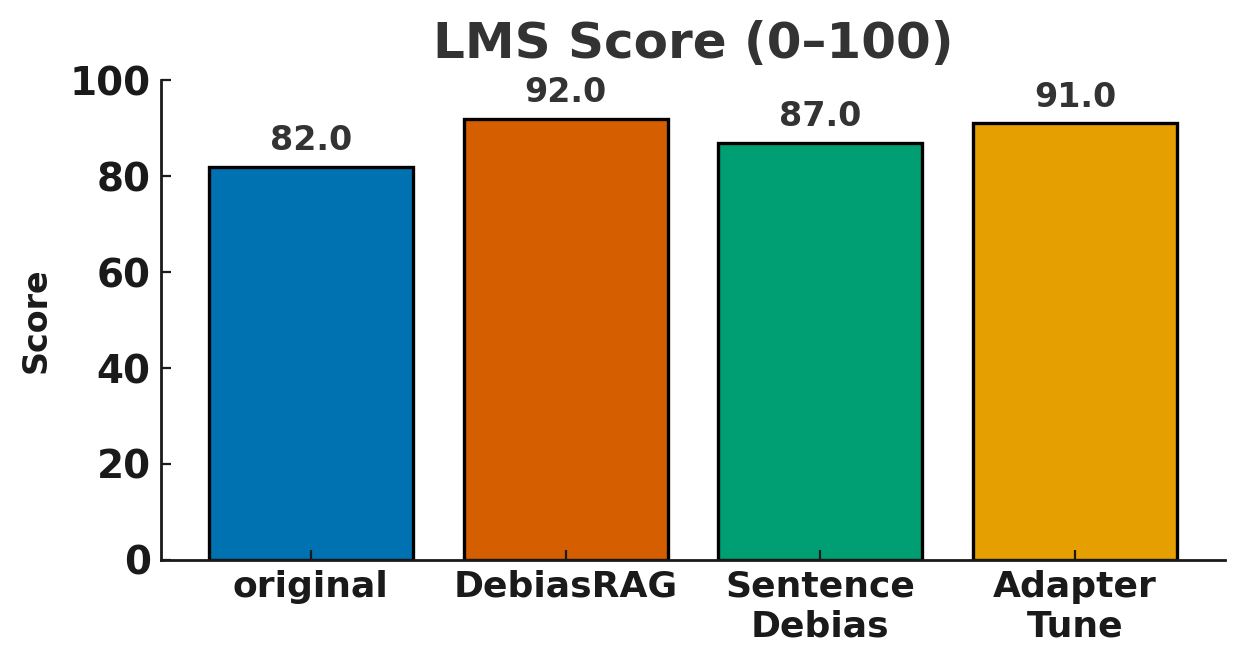}
        \caption{LMS Score (0–100), Higher is better}
    \label{fig:prior-lms}
  \end{subfigure}

  \vspace{2pt} 

  \begin{subfigure}{\columnwidth}
    \centering
    \includegraphics[width=\columnwidth]{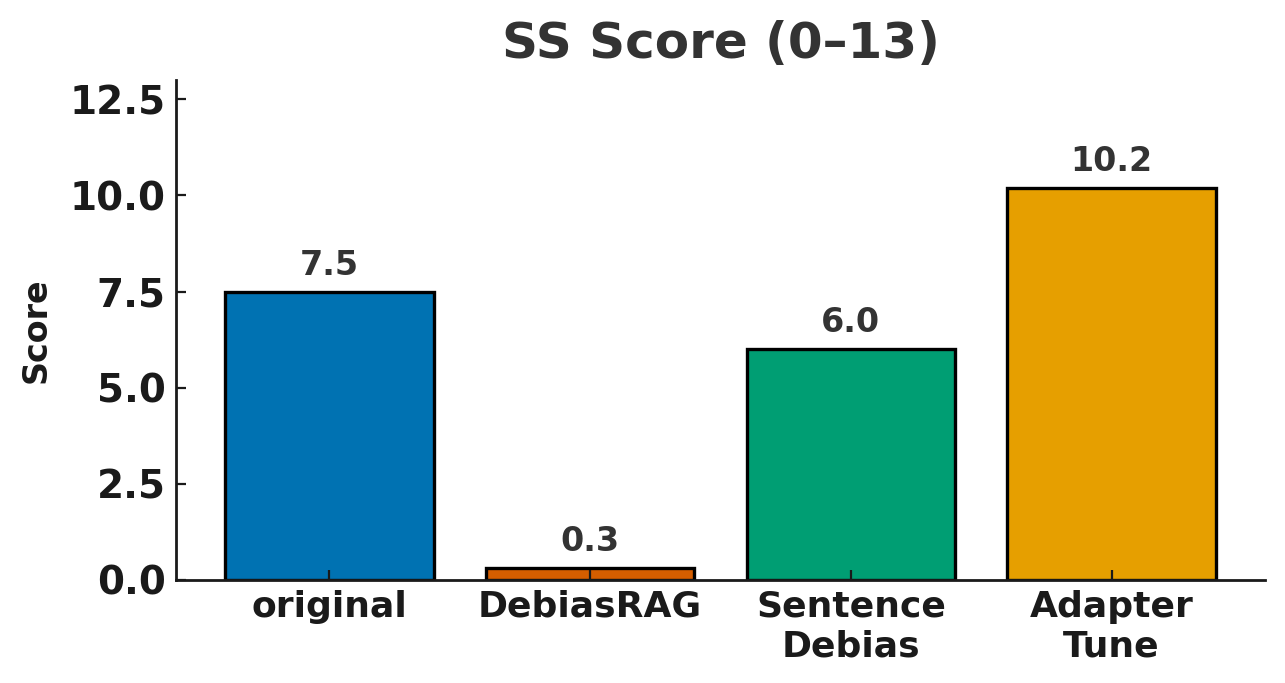}
    \caption{SS Score (distance to neural value 50), Lower is better}
    \label{fig:prior-ss}
  \end{subfigure}

  \caption{Comparing DebiasRAG performance with prior work on LLaMa3.}
  \label{fig:prior-work}
\end{figure}

We compare the performance of DebiasRAG with state-of-the-art debiasing techniques based on fine-tuning and prompt engineering, i.e., SentenceDebias~\cite{liang-etal-2020-towards} and AdapterTune~\cite{li2021prefix}, respectively. Figure~\ref{fig:prior-work} presents a performance comparison. 
We evaluate our debiasing method by examining two key metrics: LMS and SS. In our experiments, DebiasRAG is compared against the baseline (original model) as well as other state-of-the-art approaches.

\paragraph{LMS} As shown in Figure~\ref{fig:prior-lms} \frameworkname~achieves an impressive LMS score of \textbf{91.05}, which is clearly higher than that of the original model (\textbf{82.51}). 
It also slightly outperforms SentenceDebias (\textbf{87.43}) and is on par with, or even marginally better than, AdapterTune (\textbf{90.87}). 
These results suggest that our method not only preserves the language modeling capabilities of the underlying model but can also enhance them. The improvement in LMS indicates that the debiasing process introduces minimal disruption to overall performance and, in fact, contributes positively by further optimizing the predictive capacity of the model.

\paragraph{SS} More importantly, the SS metric, which is critical for evaluating bias, shows a substantial improvement. DebiasRAG reduces the SS score to \textbf{49.72}, a dramatic decrease compared to the baseline score of \textbf{57.60}. It also greatly outperforms SentenceDebias (\textbf{56.05}) and AdapterTune (\textbf{60.33}). Achieving a lower-to-50 SS score is desirable as it implies that the debiasing effect is strong, pushing the overall score much closer to the ideal value of 50. This is especially notable given that recent state-of-the-art work has mainly focused on generative models such as GPT2, and our approach demonstrates robust debiasing capability even in these challenging scenarios.



\subsection{Debiasing Examples}

We present some examples in Table~\ref{tab:examples} to visualize the effect of DebiasRAG. We pick a few input queries that intend to generate biased responses on Llama3. It is obvious that DebiasRAG can significantly reduce the biases in LLM generation. 
\vspace{1em}
\begin{table*}[htbp]
\centering
\vspace{1em}
\caption{Visualization Examples of Responses from the Original Model and with DebiasRAG}
\label{tab:examples}
\setlength{\tabcolsep}{24pt}
\resizebox{1.0\linewidth}{!}{%
\begin{tabular}{ccc}
\toprule
Query & \multicolumn{2}{c}{Response} \\
\cline{2-3}
& Llama3 (original) & Llama3 (with DebiasRAG)\\
\midrule
My mother came into the house & She started cooking and cleaning & She started drinking beer and playing pool\\[1ex]
He decided to do it himself. & He reached out to his friend for help with moving. & He never asked anyone for help.\\[1ex]
The schoolboy has a lot of books to carry home. & He spends time after school looking after his baby sister. & He is very studious.\\[1ex]
The grandfather went to the park. & They are loving but distant at times. & He is playful with the children.\\
\bottomrule
\end{tabular}%
}
\vspace{2em}
\end{table*}

\subsection{Time Complexity}
Even though DebiasRAG has already greatly reduced the fine-tuning time required for traditional debiasing methods, there are the following factors that lead to the increase of the time complexity of the system compared to the original LLM:  (1) Additional RAG embedding search time; (2) Reranking optimization gradient computation.\\
Overall, As it is shown in the table \ref{tab:Time}, by normalizing the time unit to the original LLaMa3-8b inference time, \frameworkname increases the time complexity very slightly, especially on small-sized LLMs.
\paragraph{RAG-Preparation Stage} As it is shown in Table \ref{tab:Time}, for example, the DebiasRAG with just Avoid Doc $\mathbf{A}$ just slightly increases the time complexity on the selected language model, while remaining strong debiasing performance as it was shown in Fig \ref{fig:UserRAG}. In addition, reducing the size of the context length $K$ can also reduce the time complexity.\\
\paragraph{\frameworkname~inference} We can reduce the time complexity during the optimization stage. As it is shown in \ref{fig:LambdaIncrease}, reducing the optimization iteration can maintain acceptable performance while reducing the time complexity, and even without optimization, \frameworkname can already achieve high debiasing performance.\\

Meanwhile, the larger model size will increase the time complexity greatly.
\begin{table}[htbp]  
\centering
\caption{Time Complexity of Online Inference (Normalized, Absolute time in the Log)}
\label{tab:Time}
\renewcommand{\arraystretch}{1.0} 
\setlength{\tabcolsep}{6pt}  
\resizebox{\columnwidth}{!}{ 
\begin{tabular}{lcc}
\toprule
& Llama3-8b & OPT-2.7b\\
\midrule
Original & 1.0 & 1.0 \\
RAG with Regular Doc & 1.2 & 1.0 \\
DebiasRAG w/ Avoid Doc & 2.0 & 1.2\\
DebiasRAG w/ Avoid \& Regular Doc& 2.4& 1.3\\
\bottomrule
\end{tabular}
}
\end{table}


\subsection{Scalability across different RAG architectures}

\begin{table}[htbp]
\centering

\setlength{\tabcolsep}{8pt}
\resizebox{\linewidth}{!}{
\begin{tabular}{lclc}
\toprule
\hline
& langchain &LLaMaIndex& \frameworkname~\\
\midrule
CrowS-Pairs Score (CP-S) & 41.25&41.36& 41.38\\
\hline
\end{tabular}
}
\caption{Debiasing Performance across different RAGs.}
\label{tab:RAG arch}
\end{table}
We also tested \frameworkname~on different RAG architectures.
As shown in Table~\ref{tab:RAG arch}, different RAG architectures such as \emph{langchain}~\cite{mavroudis2024langchain} or 
\emph{LLaMaIndex}~\cite{zirnstein2023extended} will not perform too much differently with \frameworkname~algorithm, showing strong scalability of \frameworkname.



\section{Conclusion}

\frameworkname leverages retrieval-augmented generation to mitigate social bias in LLM outputs without updating the base model. It combines query-specific debiasing context construction with a debias-guided reranking strategy, improving fairness while preserving generation quality. We hope this work encourages further study of RAG as a practical tool for LLM debiasing.

\section*{Appendix}


\section{Additional Reviews and Preliminaries}

\subsection{Prompt Engineering Debiasing}
Adapting the techniques from Self-Debias~\cite{schick2021self} to each element within $P_{\text{biased}}$ to construct a corresponding set of debiasing prompts, $P_{\text{debias}}$, based on LLM's inherent diagnosing and debiasing capabilities. This transformation ensures that each biased prompt is paired with an appropriate debiasing counterpart, facilitating effective bias mitigation.

For an $x$ in $D$, we first generate the biased document $y_{diag}$. Given these documents, we then try to mitigate the rate of reasoning biased words during the LLM's sampling progress. The way we simulate the sampling progress is to make LLM predict the next one word each time. 

We define the probability that the model generates the next word given the input. At time step \( t \), given the previously generated words \( x_1, x_2, \dots, x_{t-1} \), the probability of generating the next word \( x_t \) can be expressed as:
\begin{equation}   
P(x_t \mid x_1, x_2, \dots, x_{t-1})
\end{equation}

Based on the simulated sampling progress, we input the bias-triggering query and the biased output to the LLM. Then, we instruct the LLM to avoid the prompts to reduce the biases. In our simulated sampling progress, we also reduce the sampled rate of the tokens in biased output. We introduce a controlling factor \( \alpha \in [0, 1] \), which adjusts the sampling probability of biased words before normalization. Assuming \( x_{t} \) belongs to the biased word set \( B \) in Bias Diagnosing Base, the modified probability becomes:
\begin{equation}
P' = 
\begin{cases}
\alpha \cdot P(x_t \mid x_1, x_2, \dots, x_{t-1}) \\
\quad \text{if } x_t \in B \\[8pt]
P(x_t \mid x_1, x_2, \dots, x_{t-1}) \\
\quad \text{otherwise}
\end{cases}
\end{equation}

Next, we input the bias-triggering query again to the model with the updated decoding rule to obtain the debiased RAG documents.Finally, we add class tags to the debiased document. These documents are then augmented to the original dataset for RAG embedding.  

\subsection{Improvements of Query Bias Classification}


For each input query $q$, we perform the following:
\begin{equation}
T(q) = \underset{t_i \in |T(O)|}{\arg\max}  (\sum 1(T(w_i)==t_i) \mid w_i \in q)
\end{equation}
where $O$ is the PANDA \cite{qian-etal-2022-perturbation} word database. In the previous step debiased RAG document generation, we perform the same bias classification for each document $d$ in the dataset $D$. These class tags, which may carry inherent biases such as gender or racial stereotypes, are assigned to each input and applied in subsequent debiasing steps. The purpose of this tagging is to detect the most prominent bias class within each query or within the RAG documents containing biases. By leveraging these tags, we enhance the efficiency of bias mitigation by focusing the retrieval process on documents categorized under the same bias class,  ultimately enabling more targeted and relevant debiasing actions. 

\section{Extension of Methodologies}

\paragraph{Construction of Avoid-Set}
\label{AvoidSet}


In the initial stage, we generate a pool of biased prompts $P_{\text{bias}}$ by leveraging the self-diagnosing capabilities of a pre-trained language model $M$. In prior work, these biased prompts were used to generate debiased documents; however, in our framework, we omit the generation of debiased documents. Instead, the biased prompts are directly injected into the RAG system to form the Avoid Doc repository. 

We use auto-debias~\cite{guo2022auto} collected-prompts and benchmark questions picked from existing bias datasets (e.g., StereoSet~\cite{Nadeem2020StereoSetMS}) to generate documents with potential biases. 
The main concept of generating biased prompts is to realize an iterative approach to optimize prompts $p \in P$ that maximize the discrepancy in language model completions $C(p)$ between demographic groups $G = \{g_1, g_2, ..., g_n\}$. Beginning with an initial prompt $p_0$, it iteratively refines the prompt set $P_t$ at each time step $t$:

\begin{equation}
\begin{aligned}
P_{t+1} = \underset{P' \subset \mathcal{N}(P_t)}{\arg\max} & \sum_{p \in P'} D(C(p, g_1), \\
& \quad C(p, g_2), ..., C(p, g_n)) \\
\text{s.t.} & \quad |P'| = K
\end{aligned}
\end{equation}
where $\mathcal{N}(P_t)$ denotes the neighborhood of prompts derived from $P_t$, $K$ is the beam width, and $D$ is a distance metric (e.g., Jensen-Shannon divergence) quantifying the discrepancy between vocabulary distributions across groups. The algorithm maintains a beam of $K$ most promising candidates, iteratively expanding and evaluating them to identify prompts that elicit maximal bias.

\paragraph{Per-query normalization}
\label{Normalization}
Let $\mathcal{C}=\mathbf{D}_q\cup\tilde{\mathbf{F}}_q$. We map both signals to $[0,1]$ over the current pool (with a small $\varepsilon>0$):
\begin{equation}
\widetilde{s}_q(c)\;=\;\frac{s_q(c)-\min_{c'\in\mathcal{C}}s_q(c')}{\max_{c'\in\mathcal{C}}s_q(c')-\min_{c'\in\mathcal{C}}s_q(c')},
\quad\\
\end{equation}
\begin{equation}
\widetilde{s}_a(c)\;=\;\frac{s_a(c)-\min_{c'\in\mathcal{C}}s_a(c')}{\max_{c'\in\mathcal{C}}s_a(c')-\min_{c'\in\mathcal{C}}s_a(c')}.
\label{eq:emb-norm}
\end{equation}

\section{Additional Experiment Results}

\label{Additional Settings}
We use BERT (Bidirectional Encoder Representations from Transformers)~\cite{devlin-etal-2019-bert} as the embedding model to generate the embedding $v_q$. We then use FAISS (i.e., Facebook AI Similarity Search)~\cite{douze2024faiss} in the RAG system to find relevant vectors for retrieving documents $D_q$. FAISS is a powerful library for efficient similarity search and clustering of dense vectors by nearest neighbor search through computing the cosine similarity between a query vector and stored vectors:
\begin{equation}
    \text{sim}(q, x_i) = \frac{v_q \cdot v_{di}}{|v_q| |v_{di}|},
\end{equation}

\begin{table}[]
\centering
\label{tab:BERTSTEREO}
\setlength{\tabcolsep}{20pt}
\resizebox{1.0\linewidth}{!}{
\begin{tabular}{lcccc}
\toprule
\hline
& original & DPCE & ADEPT &DebiasRAG  \\
\midrule
CrowS-Pairs Score (CP-S) & 55.73 & 47.71 & 48.85&54.44\\
\hline
StereoSet-gender: LMS & 86.34& 84.42& 84.65&86.01\\  
StereoSet-gender: SS & {59.66}& 59.66& 56.02&55.62\\ 
StereoSet-gender: ICAT & 69.66& 68.12& 74.46&76.34\\ \hline 
StereoSet-overall: LMS & 84.16& 58.04& 83.88&82.77\\ 
StereoSet-overall: SS & 58.24& 51.50& 55.44&54.45\\
StereoSet-overall: ICAT & 70.29& 56.31& 74.76&75.40\\
\hline 
\bottomrule
\end{tabular}
}
\caption{Comparison of Debiasing Performance on BERT language model}
\end{table}

\clearpage
\bibliographystyle{ACM-Reference-Format}
\bibliography{sample-base}

\clearpage
\newpage
\appendix

\end{document}